\pdfoutput=1
\documentclass[11pt]{article}

\usepackage[final]{acl}

\usepackage{times}
\usepackage{latexsym}
\usepackage{url}

\usepackage[T1]{fontenc}
\usepackage[utf8]{inputenc}

\usepackage{microtype}

\usepackage{inconsolata}

\usepackage{graphicx}

\usepackage{times}
\usepackage{booktabs}
\usepackage{latexsym}
\usepackage{algorithm}
\usepackage{multirow, multicol}

\usepackage{bbm}
\usepackage{algpseudocode}
\usepackage{amsmath}
\usepackage{soul}
\usepackage{pifont}
\usepackage{paralist}
\usepackage[most]{tcolorbox}
\definecolor{green1}{RGB}{215, 255, 215}  % Medium light green
\definecolor{green2}{RGB}{138, 209, 138}  % Medium green

\algnewcommand\algorithmicinput{\textbf{Input:}}
\algnewcommand\INPUT{\item[\algorithmicinput]}

\algnewcommand\algorithmicrepresentation{\textbf{Representation:}}
\algnewcommand\REPRESENTATION{\item[\algorithmicrepresentation]}

\algnewcommand\algorithmicoutput{\textbf{Output:}}
\algnewcommand\OUTPUT{\item[\algorithmicoutput]}

\usepackage{tikz}
\newcommand*\circled[1]{\tikz[baseline=(char.base)]{
        \node[shape=circle,draw,inner sep=1pt] (char) {#1};}}

\usepackage{graphicx}
\usepackage{amssymb}
\usepackage{inconsolata}

\title{SMAB: MAB based word Sensitivity Estimation Framework and its Applications in Adversarial Text Generation}

\author{
\textbf{Saurabh Kumar Pandey\textsuperscript{1*}},
  \textbf{Sachin Vashistha\textsuperscript{2*}},
  \textbf{Debrup Das\textsuperscript{3}},
\\
  \textbf{Somak Aditya\textsuperscript{2}},
  \textbf{Monojit Choudhury\textsuperscript{1}}
  \\ 
  \texttt{saurabh2000.iitkgp@gmail.com, sachinvashistha6916@gmail.com,}
\\
  \texttt{saditya@cse.iitkgp.ac.in, monojit.choudhury@mbzuai.ac.ae}
   \\
  \textsuperscript{1}MBZUAI, 
  \\
  \textsuperscript{2}Indian Institute of Technology, Kharagpur
  \\
  \textsuperscript{3}University of Massachusetts Amherst 
}

\begin{document}
\maketitle

\setlength{\abovedisplayskip}{1pt}
\setlength{\belowdisplayskip}{1pt}

\begin{abstract}
To understand the complexity of sequence classification tasks, \citet{hahn-etal-2021-sensitivity} proposed sensitivity as the number of disjoint subsets of the input sequence that can each be individually changed to change the output. Though effective, calculating sensitivity at scale using this framework is costly because of exponential time complexity. Therefore, we introduce a \textbf{S}ensitivity-based \textbf{M}ulti \textbf{A}rmed \textbf{B}andit framework (\textbf{SMAB}), which provides a scalable approach for calculating \textit{word}-level \textit{local} (sentence-level) and \textit{global} (aggregated) sensitivities concerning an underlying text classifier for any dataset. We establish the effectiveness of our approach through various applications. We perform a case study on \textsc{CheckList} generated sentiment analysis dataset where we show that our algorithm indeed captures intuitively high and low-sensitive words. Through experiments on multiple tasks and languages, we show that sensitivity can serve as a proxy for accuracy in the absence of gold data. Lastly, we show that guiding perturbation prompts using sensitivity values in adversarial example generation improves attack success rate by 13.61\%, whereas using sensitivity as an additional reward in adversarial paraphrase generation gives a 12.00\% improvement over SOTA approaches. \textcolor{red}{\textit{Warning: Contains potentially offensive content.}}
\let\thefootnote\relax\footnotetext[1]{* indicates equal contribution, Order chosen at random}
\end{abstract}

% \iftaclpubformat

\section{Introduction}
% ------------- Figure -----------------%
\begin{figure*}[t]
\includegraphics[width=\textwidth]{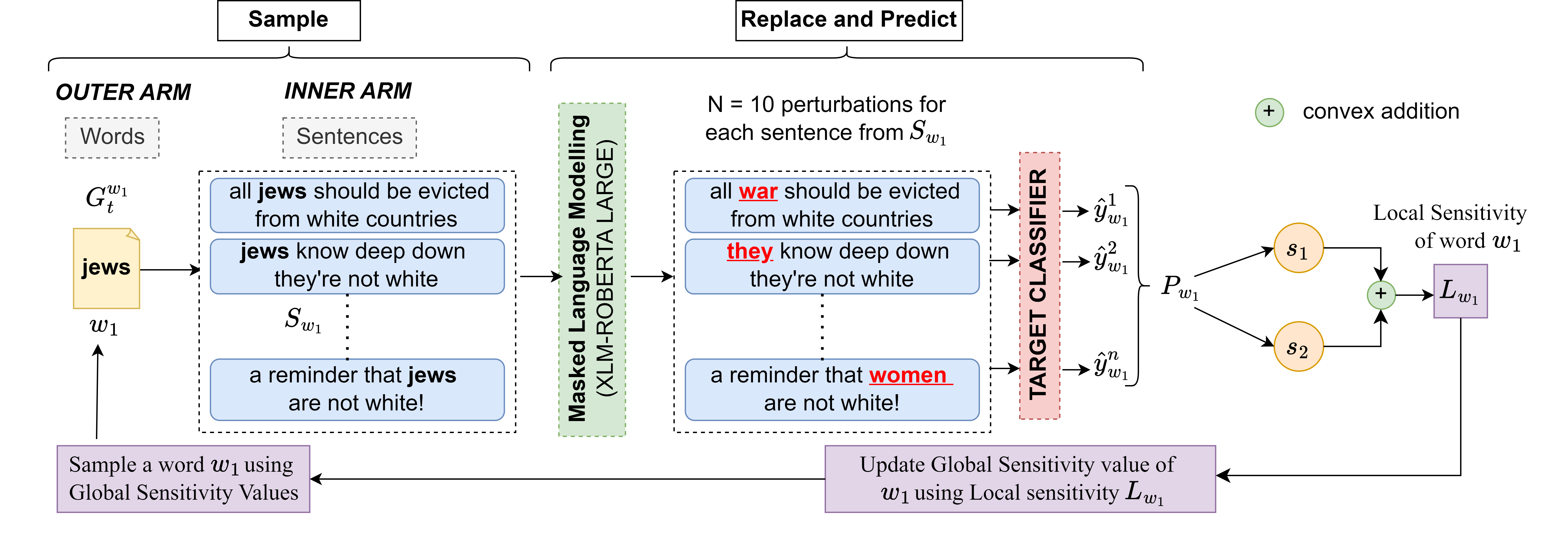}
\caption{Overview of our \textbf{SMAB} framework. The outer arm consists of all words in the corpus, each linked to a set of sentences in the inner arm. $w_1$ is a word in the outer arm, and $S_{w_1}$ is the set of sentences in the inner arm that contains $w_1$. $G_t^{w_1}$ is the Global Sensitivity of word $w_1$ at step $t$. We utilize a \textbf{sample-replace-predict} strategy to estimate local sensitivity values $L_{w_1}$ for a word $w_1$. Here, $P_{w_1}$ is the set of predicted labels obtained after perturbing $S_{w_1}$ sentences and using the target classifier. $s_1$ is chosen randomly from $P_{w_1}$ while $s_2$ is chosen such that it has the highest reward. The local sensitivity values of a word help to update its Global Sensitivity values, which helps in better outer arm selection in the next time step.
}
\label{fig:MABframework}
\end{figure*}
% ------------- Figure -----------------%

Classifiers leveraging pre-trained Language Models (\textit{PLMs}) while being empirically successful, are often opaque. Identifying input subspaces where the models classify correctly (or incorrectly) or the input patterns the model is \textit{sensitive} towards, is not straightforward. In the presence of model weights (white box), it may be feasible to explore such spaces but computationally intractable. On the other hand, methods attempting to explain black-box models resort to robust diagnostic tests (such as \textsc{CheckList}; \citet{ribeiro-etal-2020-beyond}) at scale, which requires human input and may not be sufficient. Visualization techniques such as LIME \cite{ribeiro2016whyitrustyou}, SHAP \cite{lundberg2017unifiedapproachinterpretingmodel} attempt to explain \textit{local} causes towards a prediction but do not provide a global view of these models.
Therefore, a systematic framework is required to understand the model's weaknesses and strengths related to the input space without assuming \textit{access to model weights}. 

\citet{hahn-etal-2021-sensitivity} proposed a theoretical framework for understanding the complexity of sequence classification tasks using {\em sensitivity}, where sensitivity is the number of disjoint subsets of the input sequence that can each be individually changed to change the output. High-sensitivity functions are complex because a single change in the input can alter the output, whereas low-sensitivity functions are simple. Sensitivity helps predict the complexity of the tasks for various machine learning methods. While the proposed method effectively captures the task complexity, it requires iterating exhaustively over all possible independent subsequences in the input, resulting in a time complexity that is exponential to the number of input tokens. 
\textit{In this work, we extend this definition of sensitivity} and \textit{propose a scalable framework based on multi-armed bandits to calculate word-level sensitivities on sequence classification tasks, without assuming access to model weights and gold labels}. 

Specifically, our framework provides an effective tool for \textit{local} attributions of the model output to individual segments of the input (sentence level \textit{local sensitivities}) as well as dataset-level sensitivities to specific words/phrases, which is independent of any context (\textit{global sensitivities}). We introduce \textbf{SMAB}, a sensitivity-based multi-armed bandit framework, which utilizes masked language modeling \textit{(MLM)} to compute word-level sensitivities in the dataset through effective exploration-exploitation strategies in a multi-armed bandit setup. Using these word-level sensitivities, we prove the effectiveness of the SMAB framework on three different tasks for three different application scenarios: (1) \textbf{Case Study on \textsc{CheckList}}, where we show that for template generated datasets, the global sensitivity values obtained from SMAB can help us identify high and low sensitive words across different test types. (2) \textbf{Sensitivity as a Proxy for Accuracy}, we show that sensitivity can serve as an unsupervised proxy for accuracy for a given classifier when the gold labels are unavailable. (3) \textbf{Adversarial Example Generation}, where we propose perturbation instructions that use global sensitivities obtained from the SMAB framework along with perturbation instructions proposed in \citet{xu2023llmfoolitselfpromptbased} to attack LLMs like GPT-3.5 \cite{brown2020languagemodelsfewshotlearners} and Llama-2-7B \cite{touvron2023llama2openfoundation}. We also demonstrate that using local sensitivities as an additional reward helps design highly accurate paraphrase attacks for LMs using the adversarial attack generation method proposed in \citet{roth2024constraintenforcingrewardadversarialattacks}. The two primary  contributions\footnote{Code: \href{https://github.com/skp1999/SMAB}{https://github.com/skp1999/SMAB}} of this work are summarized as follows.
\begin{compactitem}
    \item We propose an efficient, scalable algorithm, SMAB, to estimate the \textit{local} (sentence\-level) and \textit{global} (dataset\-level) sensitivities of words concerning an underlying classifier without \textit{access to model weights and gold labels}.
    % \item We propose and empirically establish the usefulness of word sensitivity in three different applications across several classification tasks, proving both the generalizability and effectiveness of our approach.
    \item We propose and empirically establish the usefulness of SMAB by (i) a case study on a templated dataset (using \textsc{CheckList}) to identify high and low-sensitive words, (ii) proposing sensitivity as an unsupervised proxy for accuracy (drops), and (iii) adversarial example generation using \textit{local} (sentence-level) and \textit{global} (dataset-level) word sensitivities. 
    % \item Through a case study on the CheckList-templated dataset, we show that the sensitivity estimates are well-suited to identify globally high and low-sensitive words.
    % \item We further show that sensitivity can serve as an unsupervised proxy for accuracy in the absence of gold labels
    % \item We propose perturbation instructions that include word-level sensitivities to attack LLMs in a zero-shot manner. We also show that local sensitivities obtained from SMAB can be used as an additional reward to generate adversarial examples through finetuning.
   % \item We propose perturbation prompts that include words with high sensitivity values obtained from SMAB and also show that, when used as an additional reward, sensitivities of a word can boost the adversarial attack success rate of a known attack method.
\end{compactitem}

\section{Methodology}
We formalize the definitions of the global sensitivity of a word for a given text classifier. Subsequently, we explain our proposed sensitivity estimation framework with examples.

\subsection{Problem Formulation}
Given an input space $X$ containing the input sentences and output space of possible labels $Y$, we have a pre-trained classifier \(f_\theta: X \rightarrow Y \) that maps the input text $ x = [w_1 \cdot w_2 \cdot w_3 \cdots w_n]  \in  X $ to a class $\hat{y} \in Y$. We are interested in finding the minimal subset of words to replace in $x$ (by contextually relevant words), such that for the new sentence $x'$, $f_\theta(x') \neq f_\theta(x)$.

\subsection{Definitions}
\label{sec:definitions}
\textbf{Local Sensitivity}:~~  We define the local sensitivity of a word for a specific input text ($x$). For an underlying classifier ($f_\theta$), local sensitivity estimates the relative importance of a word towards the predicted label ($f_\theta(x)$). We estimate singleton sensitivity \cite{hahn-etal-2021-sensitivity}\footnote{For subset sensitivity, \citet{hahn-etal-2021-sensitivity} captures the variance among predicted labels, we capture mean over flips of the predicted label from the original label.}, which is proportional to the number of flips of the predicted label when we replace a target word, say $w_i$, with contextually relevant words. \\
\\
\textbf{Global Sensitivity}: {\it Consider a text containing \(m\) words $W_x = \{w_1, w_2, \ldots w_m \}$. For an underlying classifier ($f_\theta$), we assume there exists a minimal subset of words $W_k \subseteq W_x$ that can be replaced to change the predicted label. The global sensitivity of a word provides a greedy heuristic to discover such a minimal subset. The higher the global sensitivity, the higher the chance that the word belongs to the minimal subset.} We estimate the global sensitivity of a word by aggregating the local sensitivity of words per sentence.  

\subsection{SMAB Framework}
\label{sec:SMAB}
Multi-armed bandits offer a simple yet powerful framework for algorithms to optimize decision-making over a given period of events. Our proposed framework SMAB can interpret the importance of all words (\textit{global sensitivity}) present in a given dataset for a particular task from a language modeling perspective. The framework is described below.\\
\\
\textbf{Multi-armed Bandit.} Our use of multi-armed bandits has two levels -- with words at the outer arms and sentences in the inner arm. The \textbf{outer arm} consists of all the words present in a dataset obtained after applying preprocessing techniques like removal of stopwords, removal of random URLs, and lemmatization. 
The outer arm is associated with a reward value termed as \textit{Global Sensitivity} of a word $w$, which provides a greedy heuristic to discover a minimal subset of words in a sentence that needs to be changed to flip the predicted label. The \textbf{inner arm} comprises all the sentences from the dataset in which a particular word \(w\) from the outer arm is present. We denote ${S}^{w}$ as the set of all the sentences in which a particular
outer arm or word (say \(w\)) is present. Each inner arm also has a reward value, defined as \textit{Local Sensitivity} for the inner arm.\\
\\
\textbf{Calculating Local Sensitivity: } Local Sensitivity ${L}_w$ for a word $w$ at each step $t$ is calculated by perturbing sentences using \textit{sample-replace-predict} strategy. First, we \underline{sample} a word from all the words present in the outer arm using either \textbf{Upper Confidence Bound1 (UCB)} \cite{auer2002finite} or \textbf{Thompson Sampling (TS)} \cite{thompson1933likelihood}. We use \textbf{TS} to draw a sample with maximum value from the Beta distribution of sensitivity of all words using:
\begin{equation}
    w^*_{t+1} = \operatorname*{argmax}_{w \in W} \left( Beta(\alpha, \beta) \right),
\end{equation}
where $\alpha$ $\in$ (0,1) and $\beta = 1-\alpha$. Then, we  \underline{replace} the word $w$ in all the sentences of the inner arm using predictions of a masked language model, ensuring that the resulting sentence with the new word $w'$ remains coherent and semantically sound. This process is repeated $N$ times (here $N=10$). If, across the $N$ replacements, the new word $w'$ matches the original word $w$, we discard that instance. Let $P_w$ denote the set of all valid instances i.e., instances that have not been discarded. Lastly, we use the target model to \underline{predict} the labels of the newly constructed ${P}_w$ sentences. By utilizing the target model predictions of the ${P}_w$ sentences, we select a randomly sampled sentence $s_1$ $\in$ $P_w$ with reward $r_1$ (sentence-level local sensitivity) and a sentence $s_2$ $\in$ $P_w$ with the highest reward $r_2$ is selected. The local sensitivity for a word $L_w$ is calculated as the convex combination of rewards from $s_1$ and $s_2$,  $\epsilon$ \(\in\) (0, 1) \begin{equation}
    L_w = \epsilon\:*r_1 + (1-\epsilon)\:*r_2,
\end{equation} 
\textbf{Calculating Global Sensitivity:} Global sensitivity value (represented as $G^{w}_{t}$) for a particular word $w$ at a particular step/iteration $t$ is calculated as follows: \begin{equation}
    G^{w}_{t} = \frac{(N^w\:*\:G^{w}_{t - 1}\: + \: L_{w})}{ 1 \:+ \:N^w},
\end{equation}
where $N^w$ represents the number of times the word $w$ has been picked up till now. $G^{w}_{t}$ \(\in\) (0,1). We assign $L_{w}$ to $1$ if $L_{w}>0$. We minimize the total regret $R_t$ over the total number of iterations. The estimation of total regret and the pseudocode of the complete algorithm is presented in Algorithm~\ref{algo:SMAB}.

\subsection{SMAB Training Details}
We initialize the global sensitivity values of all the words (arms) present in a dataset with $Beta(\alpha, \beta)$, where $\alpha = Random(0,0.5)$ and $\beta = (1 - \alpha)$. We use $\epsilon$ in the convex addition function as 0.9. We iterate with the total number of steps, $N = 2,00,000$. We report the  \textit{\#datapoints} used for training, inner and outer arm details in Table~\ref{tab: SMAB_training_details}. 
%Details of the target classifier being used for each dataset/task in subsequent sections can be found in Table~\ref{tab: SMAB_training_details}

\section{SMAB: A Case Study on \textsc{CheckList}}
\paragraph{Dataset.}Evaluating the global sensitivity values of words is challenging as they depend on a specific task and classifier combination. There is no straightforward way to determine the ground truth global sensitivity of different words. Therefore, we start with a template-generated dataset such as \textsc{CheckList} \cite{ribeiro-etal-2020-beyond}, where for a template, we know that changing specific keywords may cause a label flip while changing others should not have any effect on the label. We note that this still provides only an approximation of sensitivity values, as the global sensitivity value of a word (as defined in \S\ref{sec:definitions}) also depends on the target classifier.

\paragraph{Method.}The \textsc{CheckList} framework creates targeted test cases inspired by standard software engineering practices. Each test belongs to one of the categories -- MFT (\textit{Minimum Functionality Test}), INV (\textit{Invariance Test}), and DIR (\textit{Directional Expectation Test}). INV applies perturbations that preserve the original label, whereas DIR tests if the confidence of a label changes in a specific direction. These tests (INV \& DIR) consist of templates that vary according to the test type in consideration. For example, \textit{change names} test suite of \textit{INV} test type has sentences where we only vary a single word (name) in the complete sentence. In the texts shown below, only the name (\texttt{Alicia}) changes in the newly created example.

\begin{tcolorbox}[colback=white,colframe=red!75!black,left=1pt,right=1pt,top=0pt,bottom=0pt]
\textbf{\underline{Example: }}\\
    \-@JetBlue Thank you {\textcolor{blue}{Alicia}}!Exceptional Service\\
    \-@JetBlue Thank you {\textcolor{blue}{Haley}}!Exceptional Service 
\end{tcolorbox}
We utilize test types from INV and DIR to identify words with low and high sensitivity using our \textit{SMAB} framework. We experiment with two outer arm sampling strategies, UCB and TS. 
We use the perturbed Twitter US airline sentiment dataset \footnote{https://github.com/marcotcr/checklist} obtained from \textsc{Checklist}. We sample $\sim35k$ sentences from all the test types covering \textbf{38 test types}. We extract \textbf{8498} arms from the above sentences after some initial preprocessing (stopwords removal, lemmatization). Finally, we run our SMAB algorithm (\S\ref{sec:SMAB}) on the above set of arms and sentences to obtain the global sensitivity of all the words.

\begin{table}[t!]
    \resizebox{\columnwidth}{!}{%
    \large
    \begin{tabular}{c|l|c} 
        \hline
        \textbf{\textsc{Type}} & \multicolumn{1}{c|}{\textbf{\textsc{Example}}} & 
        \textbf{$\textbf{G}_s^w$}  \\ 
        \hline
        \textbf{INV}   & \begin{tabular}[c]{@{}l@{}}@united happens every time in and out of\\\textcolor{blue}{\textbf{<Newark>}}\end{tabular} & 0.0397          \\ 
        \hline
        \textbf{INV} & \begin{tabular}[c]{@{}l@{}}@JetBlue and of course that was supposed to say\\\textcolor{blue}{\textbf{<Jeremy>}}, not login.\end{tabular} & 0.0883          \\ 
        \hline
        \textbf{DIR}   & \begin{tabular}[c]{@{}l@{}}Thanks @JetBlue. Next up we will see how\\the slog from JFK to the city goes. You are\\\textcolor{red}{\textbf{<exceptional>}}.\end{tabular} & 0.6185          \\ 
        \hline
        \textbf{DIR}  & \begin{tabular}[c]{@{}l@{}}@USAirways Delays due to faulty engine light.\\Great work guys. Coming up on 2 hrs sitting\\on the plane. WorstAirlineInAmerica. You are\\\textcolor{red}{\textbf{<creepy>}}.\end{tabular} & 0.7574          \\ 
        \hline
        \textbf{INV} & \begin{tabular}[c]{@{}l@{}}@SouthwestAir I did ....it’s just been such a\\\textcolor{red}{\textbf{disheartening}} experience for me and my family\\...and a lot of taxi money wasted. \textbf{<@MZ0ql9>}\end{tabular} & 0.9311   \\
        \hline
    \end{tabular}
    }
    \caption{A few examples from the \textsc{CheckList} test suite showing the highlighted words in the template and their respective estimated global sensitivity ($G_s^w$) using SMAB. The templated words are enclosed in <word>. The low and high sensitivity words are highlighted in \textcolor{blue}{blue} and \textcolor{red}{red} respectively. %\skp{I think we should put this instead of scatter plot}
    } 
    \label{tab:template_examples}
\end{table}

\paragraph{Observations.} We observe that the perturbation of words (arms) present in \textbf{INV} templates does not tend to change the label of the original sentence. In contrast, the words in \textbf{DIR} templates are more prone to flipping the label since they contribute to confidence score manipulation. Further analysis (details in Figure \ref{fig:scatter_plot} in the Appendix) shows that the words from \textit{DIR} templates have higher estimated global sensitivity and have a much wider spread of values ranging from 0 to 1. In contrast, as expected, the words from \textit{INV} are concentrated in the low-sensitivity range of (0-0.2).
Further, in Table \ref{tab:template_examples}, we present some qualitative examples from various test types, words, and their estimated global sensitivities.

\paragraph{Evaluation.} Sensitivity threshold is the value above which a word present in a sentence if perturbed, is highly likely to change the predicted label. To evaluate the performance of our framework, we propose a metric, \textbf{Sensitivity Attack Success Rate (SASR)}, calculated as follows. Given a test dataset, a word, $w$, from the set of all the words present in the dataset, $S_w$, the set of sentences in which the word is present,   and $G_s^w$, its estimated global sensitivity from our SMAB framework, if the word is above the sensitivity threshold and replacing the word with the predictions of a masked language model flips the predicted label in any one of the sentences from $S_w$, it is called a success. SASR is the fraction of all the words in a dataset above the sensitivity threshold that can flip the predicted label.

\begin{figure}[t]
    \includegraphics[width=\columnwidth, height=0.25\textheight]{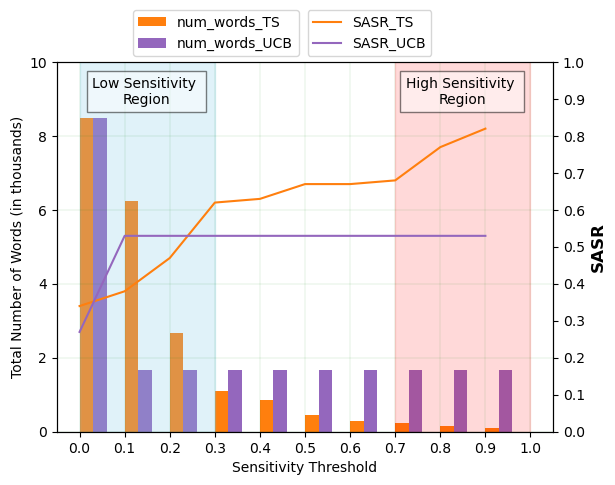}
    \caption{Variation of SASR with sensitivity threshold on CheckList test dataset for UCB and TS. For UCB, words are only present in bins (0-0.1) and (0.9-1.0), hence SASR becomes constant after 0.1. It shows that Thompson Sampling proves to be a better sampling strategy for this task as compared to UCB.}
    \label{fig:checklist_SASR}
\end{figure}

We calculate SASR for a test set from \textsc{CheckList} templated dataset of 1800 sampled data points and plot SASR for different sensitivity thresholds for both the algorithms, UCB and TS, as shown in Figure \ref{fig:checklist_SASR}. We observe that the SASR\_TS increases as we increase the sensitivity threshold, which signifies that the words in the high sensitivity region ($0.7$-$1.0$) are responsible for the change in the predicted label. SASR\_UCB remains constant after threshold of 0.1 since the estimated sensitivity values from UCB lie only in two bins, ($0.0$-$0.1$) and ($0.9$-$1.0$). Additionally, we plot the number of words obtained at different sensitivity thresholds to ensure sufficient words are present to calculate SASR. For TS, we get $104$ words even above the sensitivity threshold of $0.9$, with around $83$ of these words able to produce a flip. Hence, it shows that the estimated global sensitivities from our SMAB framework, in a true sense, capture the impact of various words in a sentence for a particular task in a given dataset.

\section{Sensitivity as a Proxy for Accuracy}
Using our multi-armed bandit framework (in \S\ref{sec:SMAB}), we calculate the word sensitivities for each word in a given dataset based on the model predictions (in an unsupervised way). Here, we aim to quantify the correlation between the accuracy of various models and the difference between their corresponding sensitivity distributions obtained from the SMAB framework. %We compare the relative drop in accuracy between the two models and the KL divergence of the sensitivity distributions from the models. 
We experiment with two different settings - (i) Correlation across languages (same model) and (ii) Correlation within language (different models). We compare the KL divergence of the sensitivity distributions from two different models (/languages) with the relative drop in accuracy between models (languages). We compute
$
\label{eq:kld}
D_{\text{KL}}(P \parallel Q) = \sum_{i=1}^{N} P(i) \log \frac{P(i)}{Q(i)}
$
, where $P$ and $Q$ represent sensitivity distributions obtained from two different runs of SMAB. $N$ represents the number of sensitivity bins (here $10$). We calculate accuracy on the same dataset using ground truth labels.
We hypothesize that KL divergence is negatively correlated with accuracy drop for both the settings, which signifies that the sensitivity distributions may serve as an \textit{unsupervised proxy for accuracy} for a given target classifier when gold labels are absent.

\subsection{Tasks \& Datasets}
\textbf{Hate Speech Classification Task.}~~~
Hate Classification is a challenging task that contains many words spanning different sensitivity bins (contains highly-sensitive target words). We selected hate speech classification datasets from various sources, covering nine languages - English, Bengali, French, German, Greek, Hindi, Italian, Spanish, and Turkish. Hereafter, we refer to this dataset as the mHate dataset.
\\\noindent
\textbf{Natural Language Inference Task.}~~~
We use the \textit{XNLI} \cite{conneau-etal-2018-xnli} dataset, a cross-lingual NLI dataset for this task, which expands upon the English-based \textit{MultiNLI} dataset \cite{williams-etal-2018-broad} by translation into 14 languages. We select five languages - English, French, Greek, Hindi, and Spanish to evaluate our hypothesis.

\subsection{Correlation Across Languages}
Robust evaluation and benchmarking of low-resource languages have always been challenging because of the lack of sufficient and reliable evaluation datasets~\cite{ahuja2022beyond}, \cite{ahuja2022multi}. SMAB may be highly effective in the evaluation and benchmarking of low-resource languages. We experiment with various languages of mHate and XNLI datasets. We quantify the relative zero-shot drop in accuracy and attempt to correlate it with the KLD. We utilize the mBERT~\cite{devlin2019bert} classifier for mHate and mDeBERTa~\cite{he2021debertadecodingenhancedbertdisentangled} for XNLI. Given a classifier, we get the predictions on the mentioned language split of the dataset. Then, we compare the KLD between sensitivity distributions of different languages from a base language (English, in our case) and the model's accuracy in various languages. We plot KLD v/s accuracy to study nine languages for a particular target classifier under study. 

\subsection{Results}
\begin{figure}[t]
    \includegraphics[width=\columnwidth]{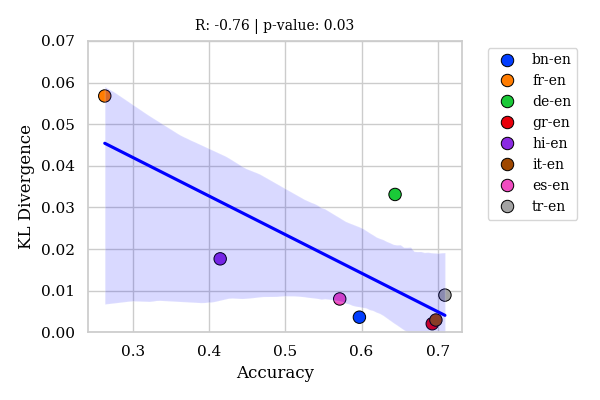}
        \caption{KL Divergence v/s accuracy across languages of mHate dataset using mBERT.}
        \label{fig:mbert_acorss_languges}
\end{figure}

\begin{figure}[t]
    \includegraphics[width=\columnwidth]{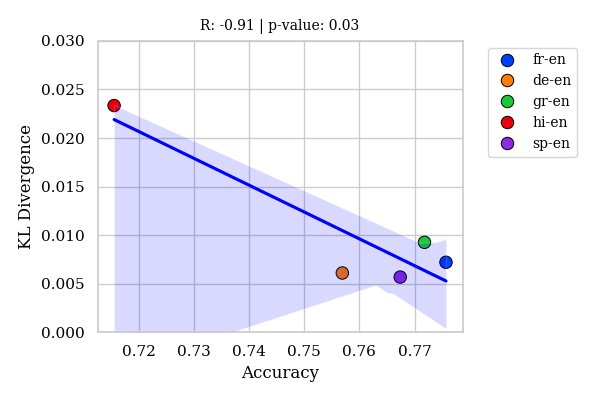}
    \caption{KL Divergence v/s accuracy across languages of XNLI dataset using mDeBERTa.}
    \label{fig:mdeberta_acorss_languges}
\end{figure}

From Figures \ref{fig:mbert_acorss_languges} and \ref{fig:mdeberta_acorss_languges}, we observe a negative correlation between KL Divergence and Accuracy on the test set. For mHate, we calculate KLD between sensitivity distributions obtained from our SMAB framework across eight languages and sensitivity distribution for English. We observe a negative correlation with Pearson Correlation Coefficient(R) of $-0.75$ (statistically significant with \textit{p-value} of $0.03$). Similarly, for XNLI, we perform it for $5$ different languages against English and obtain a Correlation Coefficient of $-0.91$ (statistically significant with \textit{p-value} of $0.03$). We also carry out experiments for \textbf{within the language} setting using various pre-trained classifiers and obtain a similar correlation (Appendix \ref{sec:appendix:subsection:kld_vs_acc}). The results show that KLD between sensitivity distributions is \textit{negatively correlated} with the accuracy of a target classifier. Hence, the sensitivity values obtained from SMAB act as an \textit{unsupervised proxy for accuracy} for a given target classifier on a dataset when the gold labels are absent.

\section{Adversarial example generation}
\subsection{Datasets} 
We evaluate the utility of word-level sensitivities in \textit{adversarial example generation} using the Sentiment Analysis task. We conduct experiments on the Cornell Movie Review dataset (RT dataset) \cite{pang-lee-2005-seeing} and SST-2 dataset~\cite{socher2013recursive}. We extend two perturbation-based (or paraphrase) attack baselines using sensitivities predicted from our SMAB framework.

\subsection{Evaluation Metrics}
\textbf{Attack Success Rate:} Following \citet{wang-etal-2018-glue}, we evaluate our attack methods using the Attack Success Rate (\textbf{ASR}). Given a dataset $\mathcal{D} = \{(x^{(i)}, y^{(i)})\}_{i=1}^{N}$ consisting of $N$ pairs of samples $x^{(i)}$ and ground truth labels $y^{(i)}$, for an adversarial attack method $A$ that generates an adversarial example $A(x)$ given an input $x$ to attack a surrogate model $f$, ASR is calculated as follows: \\
\begin{equation}
    \sum_{(x,y) \in \mathcal{D}}
    \frac{\mathbbm{1} \big[ (f(A(x)) \neq y)
                      \land (f(x) =y)\big]}
                      {\mathbbm{1} \big[ f(x) = y \big]}
    \label{eq:ASR}
\end{equation}\\
\textbf{After Attack Accuracy:} It measures the robustness of a model to an adversarial attack, A low After-attack accuracy represents a highly vulnerable model to adversarial attacks. It is calculated as follows: \\
\begin{equation}
    \frac{1}{|\mathcal{D}|} \sum_{(x,y) \in \mathcal{D}}
    \mathbbm{1} \big[ (f(A(x)) = f(x) = y) \big]
    \label{eq:AAA}
\end{equation}
% \begin{align}
%     \frac{1}{|\mathcal{D}|} \sum_{x \in \mathcal{D}} 
%     \mathbbm{1} \big[ & (L_{v}(x') = L_{g}(x) = L_{v}(x) ) \big]
%     \label{eq:AAA}
% \end{align}

%   Perturbation Instructions Table %
\begin{table*}[t!]
\resizebox{\textwidth}{!}{%
\centering
    \begin{tabular}{l | l }
    \toprule
    \textbf{\begin{tabular}[c]{@{}l@{}}Perturb\\ Type\end{tabular}} &
    \textbf{\begin{tabular}[c]{@{}c@{}}Perturbation instructions \end{tabular}} \\ 
    \midrule
    $W1$ & Replace at most two words in the sentence with synonyms.                    \\ 
    \midrule
    $W2$ & Choose at most two words in the sentence that do not contribute to the meaning of the sentence and delete them.                    \\ 
    \midrule
    $W3$ &    Add at most two semantically neutral words to the sentence.                  \\ 
    \midrule
    $W4$ (Ours) &   {\begin{tabular}[c]{@{}l@{}}   For a given sentence, there always exists a minimal subset of words that need to be replaced to flip the label of the \\sentence while preserving its semantic meaning. Global Sensitivity of a word provides a greedy heuristic to discover \\such a minimal subset. The higher the global sensitivity, the higher the chance that the word belongs to the minimal \\subset. Given the minimal subset of the words ["\texttt{[Word1]}", "\texttt{[Word2]}"] and their global sensitivity values in the \\decreasing order [\texttt{GS1}, \texttt{GS2}], replace these words in the original sentence with semantically close words.\end{tabular}}              \\
    \midrule
    $W5$ (Ours) &      {\begin{tabular}[c]{@{}l@{}}The words "\texttt{[Word1]}" and "\texttt{[Word2]}" are highly sensitive in the given sentence, and perturbing either "\texttt{[Word1]}", \\"\texttt{[Word2]}", or both can change the label of the sentence while preserving the semantic meaning of the new sentence \\as that of the original.\end{tabular}}              \\
    \midrule
    $W6$ (Ours) &      Add at most two semantically close words to the sentence, replacing the words "\texttt{[Word1]}" or "\texttt{[Word2]}, or both.\\
    \bottomrule
\end{tabular}%
}
\caption{Perturbation instructions used in Prompt Guidance: In \textit{$W4$}, \textit{$W5$} and \textit{$W6$}, \texttt{[Word1]} is replaced with the most sensitive word in the sentence, while \texttt{[Word2]} is replaced with the second most sensitive word. In \textit{$W4$}, \texttt{GS1} and \texttt{GS2} represent the global sensitivity values of the most sensitive word and the second most sensitive word. 
}
\label{tab:Attack_Guidance_perturbation_instrs}
\end{table*}

\subsection{PromptAttack using Global Sensitivities}
Our first baseline, PromptAttack \cite{xu2023llmfoolitselfpromptbased}, is a prompt-based adversarial attack that can effectively audit the adversarial robustness of a target LLM. For a target LLM, PromptAttack prompts the same LLM with a perturbation instruction such that the LLM generates an adversarial sample for a given input text, effectively fooling itself. Authors propose character-level, word-level, and sentence-level perturbations. As we capture word-level sensitivities, we define three word-level perturbation instructions utilizing the global sensitivity values obtained from the SMAB framework on the SST-2 dataset. Prompts for the sentiment classification task on SST-2 can be found in Table~\ref{tab:sentiment_classification_prompt} and all perturbation instructions are outlined in Table \ref{tab:Attack_Guidance_perturbation_instrs}. We experiment with GPT-3.5 (\texttt{gpt-35-turbo}) and Llama-2-7B as the target LLMs. Our instructions primarily rely on perturbing the high-sensitive words in the input text. We apply the \textit{fidelity filter} used in \citet{xu2023llmfoolitselfpromptbased} with the following constraints: BERTScore $\geq 0.92$ \cite{zhang2019bertscore} and Word Modification Ratio $\le 1.0$ \cite{wang-etal-2018-glue}.\\
\paragraph{Results.}As shown in Table \ref{tab:avg_ASR_ZS_SST}, using GPT-3.5 as the target LLM, we compute sensitivity values from our SMAB framework with six different LLMs serving as classifiers, referred to as \textbf{SMAB LLMs}: BERT, GPT-3.5, Llama-2-7B, Llama-2-13B, Llama-3.1-8B and Qwen-2.5-7B. For all four SMAB LLMs,  our word level perturbation instructions outperform the baselines $W1$, $W2$, and $W3$. We obtain the best ASR of \textbf{52.06\%} and the lowest After Attack Accuracy of \textbf{44.23\%} when we use Qwen-2.5-7B as the SMAB LLM. It demonstrates improvement over the baselines by a margin of \textbf{15.58\%}. Likewise, with Llama-2-7B as both the target LLM and the SMAB LLM, perturbation type $W5$ outperforms the top baseline $W3$. With Qwen-2.5-7B as both the target LLM and SMAB LLM, out perturbation type $W6$ outperforms the top baseline $W3$. The results show that our instructions utilizing high-\textit{sensitive} words consistently outperform the baselines and can generate high-quality adversarial examples as shown in Table \ref{tab:Qual_Examples_promptAttack}.

\begin{table}[!ht]
\resizebox{\columnwidth}{!}{%
    \centering
    \begin{tabular}{c|c|c|c}
    \toprule
    \textbf{\begin{tabular}[c]{@{}l@{}}SMAB \\ LLM\end{tabular}}  &
    \textbf{\begin{tabular}[c]{@{}l@{}}Perturb \\ Type \end{tabular}} & \textbf{\begin{tabular}[c]{@{}l@{}} ASR $\,\uparrow$ \end{tabular}}  & 
    \textbf{\begin{tabular}[c]{@{}l@{}} After Attack \\Accuracy $\,\downarrow$ \end{tabular}} \\
    \hline

    \multicolumn{4}{c}{\rule{0pt}{2ex} Target LLM $\rightarrow$ \textbf{GPT-3.5}}\\
    \multicolumn{4}{c}{\rule{0pt}{2ex} (Before Attack Accuracy $-$ 92.16\%)}\\
    \hline
    
    \multirow{3}{*}{\begin{tabular}[c]{@{}c@{}}\ding{55} \end{tabular}} 
     & $W1$ & 22.20 & 72.09           \\ 
    &  $W2$  & 24.63  & 69.73        \\ 
    &  $W3$  &  36.48 & 58.68   \\ 
    \hline
    
    \multirow{3}{*}{\begin{tabular}[c]{@{}c@{}}BERT \end{tabular}}  
    &   $W4$    & \cellcolor{green1} 37.32  & \cellcolor{green1}58.20   \\
    &   $W5$    & \cellcolor{green1}38.34   & \cellcolor{green1}56.96   \\
    &   $W6$    & \cellcolor{green1}48.23   & \cellcolor{green1}47.84   \\
    \hline
    
    \multirow{3}{*}{\begin{tabular}[c]{@{}c@{}}Llama-2-7B \end{tabular}}   
    & $W4$ &   \cellcolor{green1}37.49 &  \cellcolor{green1}57.85  \\
    &   $W5$  & \cellcolor{green1}38.20 & \cellcolor{green1}57.04   \\
    &   $W6$  &   \cellcolor{green1}48.26 & \cellcolor{green1}47.80  \\
    \hline
    
    \multirow{3}{*}{\begin{tabular}[c]{@{}c@{}}Llama-2-13B \end{tabular}}  
    & $W4$ &  \cellcolor{green1}38.61  & \cellcolor{green1}56.95  \\
    &   $W5$  & \cellcolor{green1}38.56  & \cellcolor{green1}56.79 \\
    &   $W6$  &  \cellcolor{green1} 50.09 & \cellcolor{green1} 46.23 \\
    \hline

    \multirow{3}{*}{\begin{tabular}[c]{@{}c@{}}Llama-3.1-8B \end{tabular}}  
    & $W4$ &   \cellcolor{green1}48.16 &  \cellcolor{green1}47.80 \\
    &   $W5$  & \cellcolor{green1}39.96  & \cellcolor{green1}55.40 \\
    &   $W6$  & \cellcolor{green1}46.98  & \cellcolor{green1}48.90 \\
    \hline

    \multirow{3}{*}{\begin{tabular}[c]{@{}c@{}}Qwen-2.5-7B \end{tabular}}  
    & $W4$ &   \cellcolor{green1}51.62 &  \cellcolor{green1}44.61 \\
    &   $W5$  &   \cellcolor{green1}38.87 & \cellcolor{green1}56.37 \\
    &   $W6$  & \cellcolor{green2}52.06 & \cellcolor{green2}44.23 \\
    \hline

    \multirow{3}{*}{\begin{tabular}[c]{@{}c@{}}GPT-3.5 \end{tabular}}
    &   $W4$    & \cellcolor{green1}37.94   & \cellcolor{green1}57.55   \\
    &   $W5$    & \cellcolor{green1}39.40   & \cellcolor{green1}56.09   \\
    &   $W6$    & \cellcolor{green1}46.06   & \cellcolor{green1}49.95   \\
    \hline

    \multicolumn{4}{c}{\rule{0pt}{2ex} Target LLM $\rightarrow$ \textbf{Llama-2-7B}}\\
    \multicolumn{4}{c}{\rule{0pt}{2ex} (Before Attack Accuracy $-$ 92.32\%)}\\
    \hline    

    \multirow{3}{*}{\begin{tabular}[c]{@{}c@{}}\ding{55} \end{tabular}} 
     & $W1$ & 56.19 & 36.13          \\ 
    &  $W2$  & 43.97 & 48.35       \\ 
    &  $W3$  & 60.20 & 32.11   \\ 
    \hline
    
    \multirow{3}{*}{\begin{tabular}[c]{@{}c@{}}Llama-2-7B \end{tabular}}   
    & $W4$ & 30.55   & 61.76    \\
    &   $W5$  &  \cellcolor{green2}61.69 & \cellcolor{green2}30.62   \\
    &   $W6$  &  44.85  & 47.46 \\ \hline

    \multicolumn{4}{c}{\rule{0pt}{2ex} Target LLM $\rightarrow$ \textbf{Qwen-2.5-7B}}\\
    \multicolumn{4}{c}{\rule{0pt}{2ex} (Before Attack Accuracy $-$ 76.38\%)}\\
    \hline

    \multirow{3}{*}{\begin{tabular}[c]{@{}c@{}}\ding{55} \end{tabular}} 
     & $W1$ & 36.70 & 39.68         \\ 
    &  $W2$  & 14.79 & 61.58       \\ 
    &  $W3$  & 28.78 &  47.71  \\ 
    \hline
    
    \multirow{3}{*}{\begin{tabular}[c]{@{}c@{}}Qwen-2.5-7B \end{tabular}}   
    & $W4$ &  28.44 &  47.94   \\
    & $W5$  & 36.01 &  40.25 \\
    & $W6$  &  \cellcolor{green2}37.16  &  \cellcolor{green2}39.22 \\

    \bottomrule
\end{tabular}%
}
\caption{PromptAttack results on SST-2 dataset. \textbf{SMAB LLM} is the LLM used for calculating sensitivity values from the SMAB framework. ASR and After Attack Accuracy are in ($\%$). All \textbf{perturb types} that outperformed the highest baseline score are highlighted in \cellcolor{green1}{green}, and the best-performing perturb type is highlighted in \cellcolor{green2}{green}.}
\label{tab:avg_ASR_ZS_SST}
\end{table}

\subsection{ParaphraseAttack using Local Sensitivities}
\label{sec:paraphraseAttack}
\citet{roth2024constraintenforcingrewardadversarialattacks} studies adversarial attacks on text classifiers using an encoder-decoder paraphrase model (T5), trained to generate adversarial examples using a reinforcement learning algorithm \cite{Williams2004SimpleSG} with a \textit{constraint-enforcing reward} that helps generation of semantically close, label invariant and grammatically correct adversarial examples. We modify the original reward function by incorporating an additional \textbf{Sensitivity reward}, defined as the difference between the sensitivity of the input text and the sensitivity of the generated text. Details of the modified reward function and the keyphrase sensitivity calculation can be found in Appendix~\ref{appendix:paraphraseattack_details}. During inference, we generate $8$ paraphrases for each input text using various decoding mechanisms under constraints used in \citet{roth2024constraintenforcingrewardadversarialattacks} and use the target classifier to predict the label. We used the RT dataset to conduct experiments using the same hyperparameter and design choices used in the referenced work. The details of the different models used, including the paraphrase model are provided in Table \ref{tab:various_models_used_paraphrase}.

\paragraph{Human Evaluation.} We also perform a human evaluation of the generated adversarial examples. Two co-authors (graduate CS students trained in NLP) quantify the quality of generated paraphrases on three aspects. We use the human evaluation metrics proposed in \citet{das-etal-2024-low}. Specificity (\textbf{SPE}) measures how specific the aspects obtained in the generated text are in response to the input. Grammaticality (\textbf{GRM}) measures how grammatically correct are the paraphrased sentences. Choose-or-Not (\textbf{CHO}) represents whether a human considers the generated paraphrase adversarial. Humans evaluate SPE and GRM on a Likert scale of $1$-$5$ (least to most). CHO is $1$ if a human considers the generated paraphrase adversarial, else $0$. We compute the average score across all instances for each metric. For human evaluation, we selected up to $100$ unique paraphrases generated by each model variation that flipped the label of a text. The number of unique paraphrases varies across strategies, as detailed in Table~\ref{tab:UniqueParaphrase_RT_appendix}. 

\paragraph{Results.} Table~\ref{tab:paraphrase_attack_RT} shows that incorporating \textit{sensitivity reward} consistently results in a higher \textbf{ASR} and \textbf{CHO} for a particular strategy and temperature combination. Each strategy uses a combination of decoding mechanisms and whether the sensitivity reward is used or not. We achieved the highest ASR of \textbf{93.03\%} with DistilBERT~\cite{sanh2019distilbert} as the SMAB Model, \textit{beam search} decoding strategy and a temperature of 1.00 that exceeds the corresponding baseline variant by \textbf{74.09\%} and the best baseline variant by \textbf{62.39\%}. Similarly, we achieved the best CHO using BERT as the SMAB model that exceeds the corresponding baseline variant by \textbf{20.14\%} and the best baseline variant by \textbf{12.00\%}. We present the qualitative examples in the Table \ref{tab:qual_examples_PP}.

\begin{table}[!htb]
\resizebox{\columnwidth}{!}{%
    \centering
    \begin{tabular}{l c c | c c c}
    \toprule
    \multirow{2}{*}{\textbf{\begin{tabular}[c]{@{}l@{}}Strategy\end{tabular}}} &
    \multirow{2}{*}{\textbf{\begin{tabular}[c]{@{}l@{}}Temp \end{tabular}}} &
    \multirow{2}{*}{\textbf{\begin{tabular}[c]{@{}l@{}}ASR \(\uparrow\) \end{tabular}}} & 
      \textbf{\begin{tabular}[c]{@{}l@{}}SPE \(\uparrow\) \end{tabular}} & 
      \textbf{\begin{tabular}[c]{@{}l@{}}GRM \(\uparrow\) \end{tabular}} & \textbf{\begin{tabular}[c]{@{}l@{}}CHO \(\uparrow\) \end{tabular}} \\ 
    & & & \multicolumn{3}{c}{\rule{0pt}{2ex} \textbf{(Human Evaluation)}}\\
    \hline

    \multicolumn{6}{c}{\rule{0pt}{2ex} SMAB Model \(\rightarrow\) \ding{55}}\\
    \hline

    DBS + (\ding{55}) & 0.85  & 30.64  & 3.43 & 3.65  & 27.00                  \\
    BS + (\ding{55}) & 0.85 & 17.27  & 3.50 & 3.85  & 22.58                      \\ 
    BS + (\ding{55})  &1.00 &  18.94  & 3.72 & 3.87  & 19.11                  \\
    BS + (\ding{55})  & 1.15 & 14.76  & 3.64 & 3.86 & 18.86                        \\ 
    \hline

    \multicolumn{6}{c}{\rule{0pt}{2ex} SMAB Model \(\rightarrow\) \textbf{BERT}}\\
    \hline
    DBS + (\ding{51})  & 0.85 & \cellcolor{green1}50.41  &  3.28&  3.35&   21.00 \\
    BS + (\ding{51}) & 0.85 & 19.77  & 3.38 & \cellcolor{green1}3.88 &  25.35 \\
    BS + (\ding{51})  & 1.00  & 22.00  &  3.46&  3.65&   24.05 \\
    BS + (\ding{51}) & 1.15 & \cellcolor{green1}50.97  & 3.52 & 3.45 &  \cellcolor{green2}39.00 \\
    \hline

    \multicolumn{6}{c}{\rule{0pt}{2ex} SMAB Model \(\rightarrow\) \textbf{DistilBERT}}\\
    \hline
    DBS + (\ding{51})   & 0.85  & \cellcolor{green1}88.85  &  \cellcolor{green2}3.88&  \cellcolor{green2}3.90&   22.00                      \\
    BS + (\ding{51})  & 0.85  & \cellcolor{green1}38.99  &  3.44&  3.57&   \cellcolor{green1}37.00                     \\ 
    BS + (\ding{51})   & 1.00  & \cellcolor{green2}93.03  &  \cellcolor{green1}3.86&  \cellcolor{green1}3.88&   26.00                      \\
    BS + (\ding{51})  & 1.15  & 20.89  &  3.40&  3.74&   22.66                     \\ 
    \bottomrule
\end{tabular}%
}
\caption{ParaphraseAttack Results on RT dataset. The strategy represents the combination of decoding mechanism (\textbf{DBS}: Diverse Beam Search, \textbf{BS}: Beam Search) and usage of sensitivity reward (\ding{55}: No sensitivity, \ding{51}: Sensitivity using BERT/DistilBERT). Temp: decoding temperature. ASR/CHO in (\%). All variants outperforming the highest baseline score are highlighted in \colorbox{green1}{green} and the best-performing variant is highlighted in \colorbox{green2}{green}.}
\label{tab:paraphrase_attack_RT}
\end{table}

\section{Related Work}
Sensitivity has been commonly used as a measure for the \textit{complexity} of sequence classification tasks~\cite{hahn-etal-2021-sensitivity}. Sensitivity has also been used for the understanding and optimization of prompts in related paradigms such as in-context learning. \citet{lu-etal-2024-prompts} performs an analysis of accuracy and sensitivity, for different prompts in an \textit{ICL} setting and observes a negative correlation between them. \textit{FormatSpread} \cite{sclar2024quantifying} also presents a Multi-armed bandit framework for a model-agnostic evaluation of performance spread across different prompt formats. 
% SMAB distinguishes itself from these works by focusing on the study and comparison of model decision boundaries through the lens of word-level global sensitivities, across all NLP tasks.
Multiple research work have focused on the paradigm of adversarial text generation by perturbing safe input examples through gradient-based approaches \cite{ebrahimi-etal-2018-hotflip,cheng2020seq2sickevaluatingrobustnesssequencetosequence}. \citet{wallace-etal-2019-universal} deployed a gradient guided search over all tokens to extract \textit{universal adversarial triggers}, which are input-agnostic tokens to trigger a model. \citet{ribeiro-etal-2018-semantically} followed a similar approach by presenting simple and universal \textit{semantically equivalent adversarial rules} (SEARs) that create adversaries on safe inputs. Other approaches such as \citet{iyyer-etal-2018-adversarial} and \citet{roth2024constraintenforcingrewardadversarialattacks}, have delved into the training of paraphrase networks for controlled generation of attack examples, whereas \citet{xu2023llmfoolitselfpromptbased} proposed prompt-based adversarial attack to audit LLMs. 

\section{Conclusion}
We introduce the notion of \textit{local} (sentence-level) and \textit{global} (word-level) sensitivities to capture the intricacies of a text classifier for a given dataset. We introduce a novel, cost-effective sensitivity estimation framework, SMAB. Through experiments on \textsc{CheckList}-generated dataset, we show that our SMAB framework captures high-sensitive and low-sensitive words effectively. We observe that the comparative accuracy between two models (for the same language or for across language on the same task) has strong negative correlations with KL divergence between (global) sensitivity distributions of the models -- showing sensitivity can be used as an unsupervised proxy for accuracy (drops). Further, we define three word-level perturbation instructions utilizing the global sensitivity values obtained from the SMAB framework to attack LLMs such as GPT-3.5 with a high success rate. We also show that sensitivity can be used as an additional reward in paraphrase-based attacks to improve the attack success rate of adversarial models. Hence, word-level sensitivities provide a closer look at how opaque language models work.

\section*{Limitations}
The work explores the proposed framework for sequence classification tasks. Further exploration is needed to extend to other tasks, such as generation and translation. The hypothesis of sensitivity acting as an unsupervised proxy is valid under the specific conditions we tested. A more detailed study of various families of models and tasks might provide deeper insights into the correlation, which will be highly useful for evaluating and benchmarking low-resource languages. It is also important to note that we performed experiments concerning adversarial example generation in English, and a full-fledged multilingual study needs to be performed.

\section*{Ethics Statement}
Although our framework helps identify words with different sensitivity levels, there can be a few repercussions. It is important to note that the method does not guarantee that the examples generated will always be adversarial. The framework, and hence the sensitivity values, may be misused by people to develop better jailbreak techniques.

\section*{Acknowledgements}
This research is partially supported by SERB SRG/2022/000648. We acknowledge the OpenAI and Azure credits from the Microsoft Accelerate Foundation Models Research (AFMR) Grant. Sachin Vashistha is partially supported by the Prime Minister's Research Fellowship (PMRF) grant.

\appendix
%You may include other additional sections here.
% \clearpage

\section{MAB Framework: Additional Details}
\label{sec:appendix_SMAB}
We summarize the details of the SMAB framework in an Algorithm format in Algorithm~\ref{algo:SMAB}. 

\paragraph{Outer Arm Selection.} We employ the UCB sampling strategy to choose a specific word $w$ at each step $t$ using:
\begin{equation*}
    w^*_{t+1} = \operatorname*{argmax}_{w \in W} \left( Beta(\alpha, \beta) \right)
\end{equation*}

% \begin{equation}
%      w^*_{t+1} = \operatorname*{arg\,max}_{w \in W} \left( \ G^w_{t} + \sqrt{\frac{2*log(1+t)}{1+N^w}} \right),
% \end{equation}
where $W$ is the set of all the words or outer arms. $N^w$ is the no. of times a word $w$ has been picked so far. 

\paragraph{Estimation of Total Regret.}~~In MAB, Total Regret $R_t$ is defined as the total loss we get by not selecting the optimal action up to the step or iteration $t$. Let the outer arm or word $w$ be picked up at the step or iteration $t$. Now, in turn, we will pick up sentences $s_1$ and $s_2$ $\in$ $S_w$. Let $L_w$ be the local sensitivity. Hence, the Total Regret $R_t$ up to the iteration $t$ is defined as:
\begin{equation}
    R_t = R_{t-1} \: + \: ([L_{w^*}\: - \: L_w] \: * \: G^w_t)
\end{equation}
where $G^w_t$ is the Global sensitivity value of (the outer arm) the word $w$ that was picked and $L_{W^*}$ is the highest value of local sensitivity that can be obtained out of the set $S^{w}$.

\paragraph{Comparison of Time Complexity.} Let $|P|$ be the size of the subset, $|D|$ be the size of the dataset i.e. the number of input sentences in the dataset, $|\sum|$ be the total number of words in the dataset, $|V|$ be the vocabulary size of the Language Model, and $cost(f)$ be the cost to use a Language Model for various purposes like classification, Masked Language Modeling.\\
\textbf{Time Complexity for Subset-sensitivity:} For every input sentence in the dataset, we use an MLM to generate $|V|^{|P|}$ perturbed strings and then classify them using an LLM to see if the label flipped. Hence, the Time Complexity is:
\begin{equation}
    O(|D|\cdot |V|^{|P|} \cdot cost(f) )
\end{equation}
For calculating the block sensitivity, the subset sensitivity is calculated K times, corresponding to the K partitions, which can go exponential. \\
\textbf{Time Complexity of SMAB:} Local sensitivity in our algorithm is closely related to the subset sensitivity defined in \citet{hahn-etal-2021-sensitivity}. Our local sensitivity is obtained by taking $|P| = 1$ (singleton sensitivity) and global sensitivity is calculated using equation 3 in our paper, which is O(1). The time complexity of our SMAB algorithm given $T$ as the total number of iterations is:
\begin{equation}
    O(T\cdot (|\sum| \, + \, (|D| \cdot |V|\cdot cost(f))))
\end{equation}
This shows that our proposed algorithm is computationally more efficient than the existing algorithms.
In practice, the top few replacements (perturbations) from the MLM probability distribution are used instead of iterating over all vocabulary symbols.

\begin{algorithm}[htbp]
    \caption{Multi-Armed Bandit Algorithm}
    \label{algo:SMAB}
    \textbf{Input: } A set of words/outer-arms \textbf{W}, Dictionary \textbf{D} containing the set $\textbf{S}^{w}$ of sentences as a \textit{value} for every \textit{key} i.e. word $w \in \textbf{W}$ and total number of iterations $\textbf{T} \gets 200000$.\\
    \textbf{Output: } The set \textbf{G} containing final global-sensitivity values for every word $w \in \textbf{W}$
    % , and the Total Regret \textbf{R}.
    \vspace{0.1cm}
    \hrule % Line after Output
    \vspace{0.1cm}

    \begin{algorithmic}[1]
        \State Initialize the set \textbf{G} as the initial values of the global sensitivities of the words. Here, $|\textbf{G}| = |\textbf{W}|$

        \State Initialize the set \textbf{N} ($|\textbf{N}| = |\textbf{W}|$) to zero. \textbf{N} represent the count of every word $w \in \textbf{W}$.

        \State $t \gets 0$
        \State \textbf{Repeat steps 5 to 9 until $t \neq \textbf{T}$ :}
        \State Select a word $w \in \textbf{W}$ such that 
        \begin{equation*}
            w^*_{t+1} = \operatorname*{argmax}_{w \in W} \left( Beta(\alpha, \beta) \right)
        \end{equation*}
    
        \State $\textbf{S}^{w} \gets \textbf{D}[w^{*}]$, $N^w \gets N^w + 1$
        \State Select two sentences \( s_1, s_2 \in \textbf{S}^{w} \) and calculate Local sensitivity as:
\begin{equation*}
    L_w = \epsilon \cdot r_1 + (1-\epsilon) \cdot r_2
\end{equation*}

        \State Update Global Sensitivity $G^w_t$ as
        \begin{equation*}
            G^{w}_{t} = \frac{(N^w \:* \:G^{w}_{t - 1} \: + \: L_{w})}{ 1 \:+ \:N^w}
        \end{equation*}

        \State Final Global Sensitivity Values: $\textbf{G}[w] \gets G^{w}_{t}$ 

        \State Total Regret $R_t = R_{t-1} \: + \: ([L_{w*}\: - \: L_{w}] \: * \: G^w_t)$
    \end{algorithmic}
\end{algorithm}

\section{\textsc{CheckList}: Additional Results}

\begin{figure}[!t]
\centering
    \includegraphics[height=0.32\textheight]{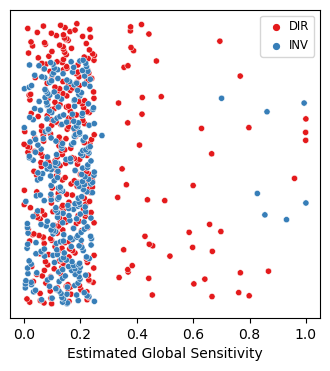}
    \caption{Scatter plot of estimated global sensitivities of arms of INV and DIR templates using TS. Words from DIR templates have higher estimated global sensitivity and are spread in the whole space as opposed to words from INV templates. %\sa{change the colors to more distinct ones. dark blue, and light red with black border.}
    } 
    \label{fig:scatter_plot}
\end{figure}

In Figure ~\ref{fig:scatter_plot}, we show a scatter plot of the word sensitivities estimated using SMAB (with Thomspon Sampling). As mentioned in the main paper, we observe that words from DIR templates have a larger distribution, present in the $0.2-1$ range, whereas, the words from the invariant test cases (INV template) have lower sensitivities (mostly between $0$ and $0.2$).

\subsection{Dataset details}
\label{sec:Appendix:subsection:dataset details}

\subsubsection{Hate Speech Dataset}
\label{sec:appendix:subsection:hate_speech_details}
In this section, we describe the sources of our compiled dataset for the hate classification task for different languages. The complete statistics for all the languages can be found in Table~\ref{tab:hate_dataset_used}.

\paragraph{English:} We used the \textbf{Stage 1: High Level Categorization} of the \textit{Implicit Hate} dataset provided by \cite{elsherief-etal-2021-latent}. It contains three labels namely implicit\_hate, explicit\_hate, and not\_hate. We converted this multi-classification task into a binary classification task where the labels implicit\_hate, and explicit\_hate are treated as the hate label.
\paragraph{Bengali:} We used a subset of the Bengali Hate speech dataset provided by \cite{romim2020hate} which includes the categories \textit{crime}, \textit{religion}, and \textit{politics} for hate label and all the categories for non-hate label.
\paragraph{Hindi:} For the Hindi language, we combined datasets collected from two different sources: 1) \cite{10.1145/3368567.3368584} provides a binary (hate/no-hate) version of the Hindi dataset. 2) We used a subset of the Hindi dataset (excluding the \textit{fake} category) provided by \cite{bhardwaj2020hostility} which includes the category \textit{non-hostile} for non-hate label and all remaining categories for hate label.

\paragraph{Spanish:}We used two datasets for the Spanish language: 1) HatEval dataset is provided by \cite{basile-etal-2019-semeval}, and 2) \cite{PereiraKohatsu2019DetectingAM} has provided the hate speech dataset in the Spanish language.

% \paragraph{Brazilian Portuguese:} We used the dataset provided by \cite{leite2020toxic} in our experiments.

\begin{table}[tbhp]
    \centering
    \begin{tabular}{c|c|c|c}
        \toprule
        \textbf{Language}   & \textbf{Hate}  & \textbf{No Hate} & \textbf{Total} \\
        \midrule
        English    &  1026 & 1658 & 2684 \\
        Bengali    & 806  & 1194 & 2000 \\
        French     & 1593 & 421 & 2014\\
        German     & 904 & 1607 & 2511\\
        Greek      & 501 & 1100 & 1601\\
        Hindi      & 783  & 549 & 1332 \\
        Spanish    & 1010  & 1290 & 2300 \\
        Turkish    & 400 & 1290 & 1690 \\
       \bottomrule
    \end{tabular}
    \caption{Dataset Statistics for val split of mHate dataset}
    \label{tab:hate_dataset_used}
\end{table}

\section{KLD v/s Accuracy Experiments}
\label{sec:appendix:subsection:kld_vs_acc}

\paragraph{Correlation Within Langauge.}
We experiment with various models on mHate and XNLI dataset. We use 5 different target classifiers for a language to estimate global sensitivities. We use XLM-R \cite{conneau2020unsupervisedcrosslingualrepresentationlearning}, mBERT, mDeBERTa \cite{he2021debertadecodingenhancedbertdisentangled}, FlanT5-L \cite{chung2022scalinginstructionfinetunedlanguagemodels} and GPT-3.5 for predictions on mHate and XNLI dataset. For a given language, we then compare the difference in sensitivity distributions of different models with respect to a base model (XLM-R), measured as KL Divergence with the performance of different models on that langauge (accuracy). We plot KLD v/s Accuracy plots for different models under study. 

\begin{figure}[htbp]
    \includegraphics[width=\columnwidth]{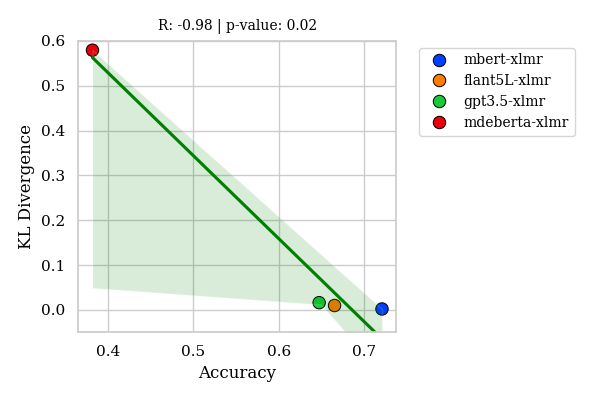}
    \caption{KLD v/s accuracy within language on mHate-English dataset.}
    \label{fig:mhate_within_languges}
\end{figure}

\section{Improving Paraphrase Attack with Local Sensitivity}
\label{appendix:paraphraseattack_details}
% In this section, we provide additional details about improving paraphrase attacks using local sensitivity. In Table~\ref{tab:paraphrase_attack_HS}, we show the flip success rates for different variants on the hate speech dataset.

\subsection{Senisitivity Reward Calculation}
\label{appendix:sensitvity_reward_calc}
We adjust the original reward function $R(x, x')$ by incorporating an additional \textbf{Sensitivity reward} $S(x, x')$ weighted by the scaling constant $\alpha \in (0, 1)$ as shown:
\begin{equation}
    R(x, x') = R(x, x') + \alpha \, S(x, x').
\end{equation}
Here, $S(x, x')$ is the difference between the sensitivity of the input text $x$ and the sensitivity of the generated text 
$x'$: $S(x, x') = s(x) - s(x')$.
The sensitivity of a text is calculated as 
\begin{equation}
    s(x) = \frac{\sum_{i=1}^{m} \sum_{j=1}^{n_i} L_{s}^{ij}}{\sum_{i=1}^{m} n_i}
\end{equation}
where ${m}$ represents the total number of keyphrases, ${n_i}$ represents the number of words in the i-th keyphrase and $L_{s}^{ij}$ represents the local sensitivity of $j^{th}$ word in the $i^{th}$ keyphrase. For our experiments, we used Scaling constant $\alpha = 0.25$. We extract the keyphrases from a text using a TopicRank keyphrase extraction model using an open source toolkit \texttt{pke} \cite{boudin:2016:COLINGDEMO}.

\begin{table*}[t]
\centering
\resizebox{0.7\textwidth}{!}{%
\begin{tabular}{l|l|c}
            \toprule
            \textbf{Purpose} & \textbf{Model}  &\textbf{Threshold}\\ 
            \midrule
            Paraphraser & \texttt{prithivida/parrot\_paraphraser\_on\_T5}  & - \\ 
            \midrule
            \begin{tabular}[c]{@{}l@{}}Target (RT) \end{tabular} & \texttt{textattack/distilbert-base-uncased-rotten-tomatoes}  & - \\
            \midrule
             % \begin{tabular}[c]{@{}l@{}}Victim (HS) \end{tabular}& \texttt{bert-base-multilingual-cased}  & - \\
             % \midrule
             \begin{tabular}[c]{@{}l@{}}Sensitivity (RT) \end{tabular} & \texttt{textattack/bert-base-uncased-rotten-tomatoes}  
            & - \\ 
            \midrule
            % \begin{tabular}[c]{@{}l@{}}Sensitivity (HS) \end{tabular} & Finetuned \texttt{bert-base-multilingual-cased}  & - \\ 
            % \midrule
            \begin{tabular}[c]{@{}l@{}}Linguistic \\ Acceptability\end{tabular} & \texttt{textattack/albert-base-v2-CoLA}   &0.5\\ 
            \midrule
            \begin{tabular}[c]{@{}l@{}}Semantic \\ Consistency\end{tabular} & \texttt{sentence-transformers/paraphrase-MiniLM-L12-v2}   &0.8\\
            \midrule
            \begin{tabular}[c]{@{}l@{}}Label \\ Invariance\end{tabular} & \texttt{howey/electra-small-mnli}  &0.2\\
            \midrule
\end{tabular}%
}
\caption{Various models as target model, in sensitivity reward functions and constraints used in Paraphrase Attack using Local Sensitivity based adversarial example generation experiment. \textbf{RT}: Rotten Tomatoes dataset. }
% \textbf{HS}: Hate speech dataset.}
\label{tab:various_models_used_paraphrase}
\end{table*}

\begin{table}[htbp]
\resizebox{\columnwidth}{!}{%
    \centering
    \begin{tabular}{l c c c}
    \toprule
    \multirow{1}{*}{\textbf{\begin{tabular}[c]{@{}l@{}}Strategy\end{tabular}}} &
    \multirow{1}{*}{\textbf{\begin{tabular}[c]{@{}l@{}}Temp \end{tabular}}} & 
      \textbf{\begin{tabular}[c]{@{}l@{}}Total generated \\ paraphrases\end{tabular}} & 
      \textbf{\begin{tabular}[c]{@{}l@{}}Total used \\paraphrases \end{tabular}} \\ 
    \midrule
    
    \multicolumn{4}{c}{\rule{0pt}{2ex} SMAB Model $\rightarrow$ \ding{55}}\\
    \hline

    DBS + (\ding{55}) & 0.85  & 62 & 62\\
    BS + (\ding{55}) & 0.85 & 53 & 53\\
    BS + (\ding{55})  & 1.00 & 110 & 100\\
    BS + (\ding{55})  & 1.15 & 68 & 68\\                       
    \hline

    \multicolumn{4}{c}{\rule{0pt}{2ex} SMAB Model $\rightarrow$ \textbf{BERT}}\\
    \hline
    DBS + (\ding{51})  & 0.85 & 181 & 100\\
    BS + (\ding{51}) & 0.85 & 71 & 71\\
    BS + (\ding{51})  & 1.00 & 79 & 79\\
    BS + (\ding{51}) & 1.15 & 183 & 100\\
    \hline

    \multicolumn{4}{c}{\rule{0pt}{2ex} SMAB Model $\rightarrow$ \textbf{DistilBERT}}\\
    \hline
    DBS + (\ding{51})   & 0.85  & 319 & 100\\
    BS + (\ding{51})  & 0.85  & 140 & 100\\ 
    BS + (\ding{51})   & 1.00  & 334 & 100\\
    BS + (\ding{51})  & 1.15  & 75 & 75\\ 
    \bottomrule
\end{tabular}%
}
\caption{Total unique paraphrases generated by each model variation from a total of 359 test instances that resulted in a label flip and the total paraphrases selected for human evaluation for the \textbf{Rotten Tomatoes} dataset.}
\label{tab:UniqueParaphrase_RT_appendix}
\end{table}

% Here is the 7.2 table

\begin{table}[htbp]
\resizebox{\columnwidth}{!}{%
    \centering
    \begin{tabular}{l c c c}
    \toprule
    \multirow{1}{*}{\textbf{\begin{tabular}[c]{@{}l@{}}Strategy\end{tabular}}} &
    \multirow{1}{*}{\textbf{\begin{tabular}[c]{@{}l@{}}Temp \end{tabular}}} &
    \multirow{1}{*}{\textbf{\begin{tabular}[c]{@{}l@{}}FSR $\uparrow$ \end{tabular}}} & \textbf{\begin{tabular}[c]{@{}l@{}}After Attack \\ Accuracy $\downarrow$ \end{tabular}} \\
    \hline
    
    \multicolumn{4}{c}{\rule{0pt}{1ex} SMAB Model $\rightarrow$ \ding{55}}\\
    \hline

    DBS + (\ding{55}) & 0.85  & 24.79 & 69.35 \\
    BS + (\ding{55}) & 0.85 & 15.87  & 82.72 \\ 
    BS + (\ding{55})  &1.00 &  17.54  & 81.05 \\
    BS + (\ding{55})  & 1.15 & 13.64 & 85.23 \\  
    \hline

    \multicolumn{4}{c}{\rule{0pt}{1ex} SMAB Model $\rightarrow$ \textbf{BERT}}\\
    \hline
    DBS + (\ding{51})  & 0.85 & \cellcolor{green1}44.87  & \cellcolor{green1}49.58 \\
    BS + (\ding{51}) & 0.85 & 18.10  & 80.22 \\
    BS + (\ding{51})  & 1.00  & 20.33  & 77.99 \\
    BS + (\ding{51}) & 1.15 & \cellcolor{green1}47.62  & \cellcolor{green1}49.02 \\
    \hline

    \multicolumn{4}{c}{\rule{0pt}{1ex} SMAB Model $\rightarrow$ \textbf{DistilBERT}}\\
    \hline
    DBS + (\ding{51})   & 0.85  & \cellcolor{green1}86.35  & \cellcolor{green1}11.14 \\
    BS + (\ding{51})  & 0.85  & \cellcolor{green1}35.93  & \cellcolor{green1}61.00 \\ 
    BS + (\ding{51})   & 1.00  & \cellcolor{green2}92.20  & \cellcolor{green2}6.96 \\
    BS + (\ding{51})  & 1.15  & 20.33 & 79.10 \\
    \bottomrule
\end{tabular}%
}
\caption{ParaphraseAttack Results on Rotten Tomatoes (\textbf{RT}) dataset. The strategy represents the combination of decoding mechanism (\textbf{DBS}: Diverse Beam Search, \textbf{BS}: Beam Search) and usage of sensitivity reward (\ding{55}: No sensitivity, \ding{51}: Sensitivity using bert/distilbert). Temp: decoding temperature. FSR/After Attack Accuracy in ($\%$). \textbf{Before Attack Accuracy} is 100\%. All variants outperforming the highest baseline score are highlighted in \colorbox{green1}{green} and the best performing variant is highlighted in \colorbox{green2}{green}.}
\label{tab:paraphrase_attack_RT_appendix}
\end{table}

% SMAB table %
\begin{table*}[htbp]
\resizebox{\textwidth}{!}{
    \begin{tabular}{|c|c|l|c|c|c|c|}
    \hline
    \textbf{Task}  & \textbf{Dataset} & \multicolumn{1}{c|}{\textbf{Target Classifier}} & \textbf{Languages} & \textbf{\#datapoints} & \textbf{\#arms} & \textbf{\#edges} \\ 
    \hline
    
    \multirow{5}{*}{\begin{tabular}[c]{@{}c@{}}Sentiment \\ Analysis\end{tabular}}          & CheckList & cardiffnlp/twitter-roberta-base-sentiment-latest & english & 34486                 & 8498 & 52359 \\ 
    \cline{2-7} & RT & textattack/distilbert-base-uncased-rotten-tomatoes & english & 1066 & 5104 & 18835 \\ \cline{2-7}  & \multirow{3}{*}{\begin{tabular}[c]{@{}c@{}}SST-2 \end{tabular}} & \multirow{3}{*}{\begin{tabular}[c]{@{}l@{}}distilbert/distilbert-base-uncased-finetuned-sst-2-english \\ gpt-3.5-turbo\\ meta-llama/Llama-2-7b \end{tabular}} & english & 872  & 4088  & 8041 \\
    \cline{4-7} & & & english & 872 & 4088 & 8041 \\
    \cline{4-7} & & & english & 872 & 4088 & 8041 \\
    \hline
    
    \multirow{9}{*}{\begin{tabular}[c]{@{}c@{}}Hate \\ Classification\end{tabular}}         & \multirow{9}{*}{\begin{tabular}[c]{@{}c@{}}Multilingual\\ hate\end{tabular}} & \multirow{9}{*}{\begin{tabular}[c]{@{}l@{}}google-bert/bert-base-multilingual-uncased\\ google/flan-t5-large\\ MoritzLaurer/deberta-v3-base-zeroshot-v1\\ gpt-3.5-turbo\end{tabular}} & bengali & 2000 & 7880 & 18929 \\ 
    \cline{4-7}  &   &  & english & 2684 & 7061 & 23149 \\ 
    \cline{4-7} &  &  & french  & 2014 & 5680 & 14184 \\ 
    \cline{4-7}  &   &   & german & 2511 & 11767 & 24226\\ 
    \cline{4-7}  &  &  & greek & 1601 & 9799 & 22031 \\ 
    \cline{4-7}  &  &   & hindi  & 1332 & 7181 & 20140 \\ 
    \cline{4-7} & & & italian & 1846 & 7551 & 17876 \\ \cline{4-7}  &  &  & spanish & 2150 & 8780            & 21982 \\ 
    \cline{4-7}  & &  & turkish & 1690 & 12308 & 20631  \\ \hline
    
    \multirow{6}{*}{\begin{tabular}[c]{@{}c@{}}Natural\\ Language\\ Inference\end{tabular}} & \multirow{6}{*}{XNLI} & \multirow{6}{*}{MoritzLaurer/mDeBERTa-v3-base-mnli-xnli} & english & 5010  & 9543  & 68156 \\ \cline{4-7}  & & & french & 5010 & 12431 & 75322  \\ \cline{4-7}  & & & german & 5010 & 13099 & 61780 \\ \cline{4-7}  & & & greek  & 5010  & 15114  & 90149 \\ \cline{4-7} & & & hindi & 5010 & 9484 & 70938 \\ \cline{4-7}  &  &  & spanish & 5010 & 11994 & 69166 \\ \hline
    \end{tabular}
    }
    \caption{Training details of SMAB for each task and target classifier and the langauges. \#datapoints represent the number of sentences in the training set, \#arms represents the unique words present in the dataset after preprocessing.}
    \label{tab: SMAB_training_details}
\end{table*}

\subsection{Key Phrase sensitivity Calculation}
\label{Keyphrase_sensitivity}
We present the key-phrase sensitivity estimation process in a pseudo-code form in Algorithm \ref{algo:alg_kp_sensitivity}. This algorithm takes as input the original text $x$ and calculates its sensitivity. The algorithm has three main steps: In the \textbf{first step}, it extracts the key phrases from the input text $x$ using the Topic Rank Key phrase extraction Model $M$ and stores them in a list $K$. This first step is shown in line $6$ of the pseudo-algorithm. In the \textbf{second step}, it calculates the sensitivities of each key phrase. The algorithm iterates through all the key phrases stored in the list $K$ and does the following: \circled{1} It extracts all the words that are present in the current key phrase and stores it in the list $words$. Similarly, it maintains a global variable $Total\_words$ that will store the total number of words extracted from each Keyphrase. \circled{2} It creates a masked sentence $x_{masked}$ from the input text $x$ by replacing all the words that are present in the current key phrase with [MASK]. \circled{3} It then uses \texttt{bert-large-uncased} to generate $10$ perturbed sentences by predicting new words at the locations where [MASK] is present in $x_{masked}$. These $10$ perturbations are stored in the list $Masked\_output$. \circled{4} The algorithm then predicts the label of the input text $x$ and $10$ perturbed sentence present in the list $Masked\_output$ and stores the predictions in the variable $P rediction\_original$ and in the list $Prediction\_List$ respectively. \circled{5} Then, it iterates through all the labels in the $Prediction\_List$ and increments the variable $flips$ if the $P rediction\_original$
doesn't match the current label. 
 \circled{6} Finally, it calculates the local sensitivity of the current keyphrase as the proportion of the total flips that we got out of the total labels i.e. $flips / \texttt{len}(Prediction\_List)$ and stores it in a list $Keyphrase\_sensitivity$ at the index corresponding to the current key phrase. This second step is shown in lines $7$ to $25$ in the pseudo-algorithm. In the \textbf{third step}, the algorithm adds up all the values in the list $Keyphrase\_sensitivity$ and divides it by $Total\_words$ to get the sensitivity of the input text $x$. This third step is shown in lines $27$ to $33$ in the pseudo-algorithm.

\subsection{Qualitative analysis of the type of Attacks}
\label{sec:appendix_qual_examples_RT}
Table \ref{tab:qual_examples_PP} shows the three different types of attacks carried out by different variants when using \textbf{distilbert} as the SMAB Model. \circled{1} \textbf{Type 1} attack involves generating adversarial examples by replacing one or more words in a sentence with new words and rephrasing the original sentence. Model variant with decoding strategy \textbf{BS + (\ding{51})} and a temperature of $0.85$  has learned to carry out \textbf{Type 1} attacks. \circled{2} \textbf{Type 2} attack involves generating adversarial examples by removing one or more words from the original sentence while preserving its semantic meaning. Model variant with decoding strategy \textbf{BS + (\ding{51})} and a temperature of $1.15$  has learned to carry out \textbf{Type 2} attacks. \circled{3} \textbf{Type 3} attack involves appending various suffixes to sentences, such as adding \textit{but it's true}, or words like \textit{but why} followed by a \textit{?}. Model variant with decoding strategy \textbf{BS + (\ding{51})} and a temperature of $1$  has learned to carry out \textbf{Type 3} attacks.

% Qualitative Examples - ParaphraseAttack %
\begin{table}[htbp!]
\scriptsize 
\begin{tabular}{l l r}
\toprule
\textbf{Attack} & \textbf{Examples} & \textbf{Flip Result} \\
\midrule
\begin{tabular}[c]{@{}l@{}}Type 1\end{tabular} & \begin{tabular}[c]{@{}l@{}} \underline{Original}: smarter than its commercials make \\it seem. \\ 

\underline{Perturbed}: 
it's smarter than the commercials \\make it appear. \\

\hline
\\
\underline{Original}: it has become apparent that the \\franchise's best years are long past. \\ 
\underline{Perturbed}: it's clear that the best years of \\the franchise are long gone.\end{tabular} & \begin{tabular}[c]{@{}r@{}}\texttt{\textcolor{blue}{pos}} $\rightarrow$ \texttt{\textcolor{red}{neg}}\\ \\ \\ \\ \texttt{\textcolor{red}{neg}} $\rightarrow$ \texttt{\textcolor{blue}{pos}}\end{tabular} \\ 
\midrule
\begin{tabular}[c]{@{}l@{}}Type 2\end{tabular} & \begin{tabular}[c]{@{}l@{}} \underline{Original}: crush is so warm and fuzzy you \\might be able to forgive its mean-spirited \\second half. \\ 

\underline{Perturbed}: crush is so warm and fuzzy you \\might forgive its mean-spirited second half \\ 
\hline
\\
\underline{Original}: you can practically hear george \\orwell turning over.
\\ 

\underline{Perturbed}: You can hear george orwell \\turning over. \\ 

\end{tabular} & \begin{tabular}[c]{@{}r@{}}\texttt{\textcolor{blue}{pos}} $\rightarrow$ \texttt{\textcolor{red}{neg}}\\ \\ \\ \\ \\ \texttt{\textcolor{red}{neg}} $\rightarrow$ \texttt{\textcolor{blue}{pos}}\end{tabular} \\ 
\midrule
\begin{tabular}[c]{@{}l@{}}Type 3\end{tabular} & \begin{tabular}[c]{@{}l@{}}\underline{Original}: provides a porthole into that noble, \\trembling incoherence that defines us all. \\ 

\underline{Perturbed}: It provides a porthole into that \\noble trembling incoherence that defines us \\all. But why? \\ 
\hline
\\
\underline{Original}: this feature is about as necessary \\as a hole in the head.
\\ 

\underline{Perturbed}: This feature is about as necessary \\as a hole in the head. But it's true. \\ 

\end{tabular} & \begin{tabular}[c]{@{}r@{}}\texttt{\textcolor{blue}{pos}} $\rightarrow$ \texttt{\textcolor{red}{neg}}\\ \\ \\ \\ \\ \\ \texttt{\textcolor{red}{neg}} $\rightarrow$ \texttt{\textcolor{blue}{pos}}\end{tabular} \\
\bottomrule
\end{tabular}
\caption{Examples of successful adversarial attacks across the three different attack types when using DistilBERT as the SMAB Model. The \textbf{Flip Result} column shows label changes from \texttt{Original Label} $\rightarrow$ \texttt{New Label}, where \texttt{pos} and \texttt{neg} denote \textit{positive} and \textit{negative} classes respectively in the \textbf{Rotten Tomatoes} dataset.}
\label{tab:qual_examples_PP}
\end{table}

% Classification Prompts Table %
\begin{table*}[htbp!]
\resizebox{\textwidth}{!}{%
\centering
    \begin{tabular}{l | l }
    \toprule
    \textbf{\begin{tabular}[c]{@{}l@{}}LLM\end{tabular}} &
    \textbf{\begin{tabular}[c]{@{}l@{}}Prompt \end{tabular}} \\ 
    \midrule
    \textbf{GPT-3.5} & \begin{tabular}[c]{@{}l@{}}Given a sentence that is a movie review, your task is to assign a label based on its sentiment. Label 1 if the sentence is a \\ positive review and Label 0 if the sentence is a negative review. Remember to only provide the label.\\
    Sentence: [input\_sentence]\\
    Label: \end{tabular} \\ 
    \midrule
    \textbf{Llama-2-7B} & \begin{tabular}[c]{@{}l@{}}Please label the sentiment of the given movie review text. The sentiment label should be "positive" or "negative". \\
    Answer only a single word for the sentiment label. Do not leave answer as empty. Do not generate any extra text. \\ 
    Text: [input\_sentence] \\
    Answer: \end{tabular} \\ 
    \midrule
    \textbf{Qwen-2.5-7B} & \begin{tabular}[c]{@{}l@{}}Please label the sentiment of the given movie review text. The sentiment label should be "positive" or "negative". \\
    Answer only a single word for the sentiment label. Do not leave answer as empty. Do not generate any extra text. \\ 
    Text: [input\_sentence] \\
    Answer: \end{tabular} \\
    \bottomrule
\end{tabular}%
}
\caption{Prompts used for sentiment classification task on SST-2 dataset}
\label{tab:sentiment_classification_prompt}
\end{table*}

% Qualitative examples (PromptAttack) %
\begin{table*}[htbp]
\resizebox{\textwidth}{!}{%
\centering
    \begin{tabular}{l l l}
    \toprule
    \textbf{\begin{tabular}[c]{@{}l@{}}Perturb \\Type\end{tabular}} &
    \textbf{\begin{tabular}[c]{@{}l@{}}Qualitative \\Example \end{tabular}} &
    \textbf{\begin{tabular}[c]{@{}l@{}}Flip \\ Result \end{tabular}}\\ 
    \midrule
    $W4$ & \text{\begin{tabular}[c]{@{}l@{}} \textbf{Original Sentence:} the title not only describes its main characters, but the lazy people behind the camera as well.\\
    \textbf{New Sentence:} The title not only portrays its main protagonists, but also the laid-back crew behind the camera as well.
    \end{tabular}} & \textcolor{red}{neg} $\rightarrow$ \textcolor{blue}{pos}              \\ 
    \midrule
    $W5$ & \text{\begin{tabular}[c]{@{}l@{}} \textbf{Original Sentence:} an important movie, a reminder of the power of film to move us and to make us examine our values.\\
    \textbf{New Sentence:} an important movie, a reminder of the influence of cinema to move us and to force us to question our beliefs.
    \end{tabular}} & \textcolor{blue}{pos} $\rightarrow$ \textcolor{red}{neg}\\ 
    \midrule
    $W6$ & \text{\begin{tabular}[c]{@{}l@{}} \textbf{Original Sentence:} It's a charming and often affecting journey.\\
    \textbf{New Sentence:} It's a charming but sometimes disconcerting journey.
    \end{tabular}} & \textcolor{blue}{pos} $\rightarrow$ \textcolor{red}{neg}\\
    \bottomrule
\end{tabular}%
}
\caption{Qualitative examples
for perturbation types $W4$, $W5$, and $W6$ when using Llama-2-13B both as the as the \textsc{Smab} Model and the target Model. The \textbf{Flip Result} column shows label changes from \texttt{Original Label} $\rightarrow$ \texttt{New Label}, where \texttt{pos} and \texttt{neg} denote \textit{positive} and \textit{negative} classes respectively in the \textbf{SST-2} dataset. In $W4$, the placeholders "\texttt{[Word1]}", "\texttt{[Word2]}", \texttt{GS1}, and \texttt{GS2} are substituted with \textit{characters}, \textit{people}, 0.9865988772394045, and 0.968535617081986 respectively. In $W5$, the placeholders "\texttt{[Word1]}" and "\texttt{[Word2]}" are substituted with \textit{make} and \textit{film} respectively. In $W6$, the placeholders "\texttt{[Word1]}" and "\texttt{[Word2]}" are substituted with \textit{affecting} and \textit{often} respectively.}
\label{tab:Qual_Examples_promptAttack}
\end{table*}

% Keyphrase sensitivity algo %
\clearpage

\begin{algorithm*}[htbp]
\begin{minipage}{\textwidth}
\begin{algorithmic}[1]
\INPUT $x \gets$ Input Text
\Require $M \gets$ TopicRank keyphrase extraction model, $C \gets$ Classifier Model, $m \gets$ \texttt{bert-large-uncased} Model 
\REPRESENTATION $G(x) \gets$ Gold label of the Input Text $x$, $C(x) \gets$ Predicted label of the Input Text $x$ using the Classifier Model $C$, $m(x) \gets$ generates $K = 10$ perturbations of the input text x using the model $m$
\OUTPUT $s \gets$ Sensitivity of the Input Text

\State \textcolor{blue}{// Initialization}
\State $s \gets 0$ \Comment{\textcolor{blue}{This variable will store the sensitivity of the input text $x$.}}
\State $Total\_words \gets 0$ \Comment{\textcolor{blue}{This variable represents the total number of words across all the keyphrases.}}
\State $Keyphrase\_sensitivity \gets \{\}$ \Comment{\textcolor{blue}{A dictionary to store the sensitivity values of each keyphrase.}}

\State \textcolor{blue}{// Extract keyphrases using the TopicRank Model $M$ and store them in the list $K$.}
\State $K \gets M(x)$

\State \textcolor{blue}{// Main algorithm}
\For{$k$ \textbf{in} $K$} 
    \State $words \gets k.\texttt{split()}$ \Comment{\textcolor{blue}{Get a list of all the words in the current keyphrase.}}
    \State $Total\_words \mathrel{+}= \texttt{len}(words)$

    \For{$w$ \textbf{in} $words$}
        \State $x_{masked} \gets$ mask the word $w$ in the text $x$ using "[MASK]"
    \EndFor

    \State $Masked\_output \gets m(x_{masked})$ \Comment{\textcolor{blue}{Store all the perturbed samples in this list.}}
    \State $Prediction\_List \gets C(Masked\_output)$ \Comment{\textcolor{blue}{Store the prediction labels of all the perturbed samples in this list.}}
    \State $Prediction\_original \gets C(x)$ \Comment{\textcolor{blue}{Store the prediction label of the input text $x$.}}

    \State $flips \gets 0$ \Comment{\textcolor{blue}{A temporary variable to count the flips.}}
    \For{$labels$ \textbf{in} $Prediction\_List$}
        \If{$labels \neq Prediction\_original$}
            \State $flips \mathrel{+}= 1$
        \EndIf
    \EndFor
    \State $local\_sensitivity \gets flips / \texttt{len}(Prediction\_List)$ \Comment{\textcolor{blue}{Total flips normalized by the total number of labels.}}
    \State $Keyphrase\_sensitivity[k] \gets local\_sensitivity$
\EndFor

\State \textcolor{blue}{// Calculate the sensitivity of the input text $x$.}
\If{$Total\_words \neq 0$}
    \State $t \gets 0$ \Comment{\textcolor{blue}{A temporary variable.}}
    \For{$value$ \textbf{in} $Keyphrase\_sensitivity.\texttt{values}()$}
        \State $t \mathrel{+}= value$
    \EndFor
    \State $s \gets t / Total\_words$
\EndIf

\Return $s$
\end{algorithmic}
\end{minipage}
\caption{Algorithm to calculate the sensitivity of Input Text}
\label{algo:alg_kp_sensitivity}

\end{algorithm*}

\end{document}